\documentclass[twocolumn,showpacs,amsmath,amssymb,groupedaddress,longbibliography, aps, prx]{revtex4-2}

\usepackage{xfrac}
\usepackage{subfigure}
\usepackage{bm}
\usepackage{graphicx}
\usepackage{amsfonts}
\usepackage{amssymb}
\usepackage{amsmath}
\usepackage[svgnames]{xcolor}
\usepackage{soul} 
\usepackage[greek,french]{babel} 
\usepackage[T1]{fontenc}
\usepackage{dcolumn}
\usepackage{bm}

\begin{document}

\title{KnitCity: a machine learning-based, game-theoretical framework for  prediction assessment and seismic risk policy design.}

\author{Ad\`ele Douin\textsuperscript{*$\ddagger$}}
\author{J. P. Bruneton\textsuperscript{$\dagger$}} 
\author{Fr\'ed\'eric Lechenault\textsuperscript{*}}
\affiliation{\textsuperscript{*}Laboratoire de Physique de l'Ecole Normale Sup\'erieure, ENS, PSL Research University, CNRS, Sorbonne University, Universit\'e Paris Diderot, Sorbonne Paris Cit\'e, 75005 Paris, France}
\affiliation{\textsuperscript{$\dagger$} Laboratoire Interdisciplinaire des Energies de Demain (LIED), CNRS UMR 8236, Université Paris Cité, 75013 Paris, France}
\affiliation{\textsuperscript{$\ddagger$} Spokesperson : adele.douin@phys.ens.fr}

\begin{abstract}
Knitted fabric exhibits avalanche-like events when deformed: by analogy with eathquakes, we are interested in predicting these "knitquakes". However, as in most analogous seismic models, the peculiar statistics of the corresponding time-series severely jeopardize this endeavour, due to the time intermittence and scale-invariance of these events. But more importantly, such predictions are hard to {\it assess}: depending on the choice of what to predict, the results can be very different and not easily compared. Furthermore, forecasting models may be trained with various generic metrics which ignore some important specificities of the problem at hand, in our case seismic risk. Finally, these models often do not provide a clear strategy regarding the best way to use these predictions in practice. 
Here we introduce a framework that allows to design, evaluate and compare not only predictors but also decision-making policies: a model seismically active {\it city} subjected to the crackling dynamics observed in the mechanical response of knitted fabric. We thus proceed to study the population of KnitCity, introducing a policy through which the mayor of the town can decide to either keep people in, which in case of large events cause human loss, or evacuate the city, which costs a daily fee. The policy only relies on past seismic observations. We construct efficient policies using a reinforcement learning environment and various time-series predictors based on artificial neural networks. By inducing a physically motivated metric on the predictors, this mechanism allows quantitative assessment and comparison of their relevance in the decision-making process.       
\end{abstract}

\maketitle

A few years ago, the scientific community was somewhat taken aback by the tremendous achievements of corporate machine learning research eg. in games ~\cite{silver2017mastering, mnih2013playing, vinyals2019grandmaster} or in natural language processing ~\cite{devlin2018bert, li2016deep}. Yet the momentum impulsed by these early developments has now galvanized a growing number of fields into the use of deep learning in particular. Ranging though robotics and computer vision~\cite{ruiz2020applications}, exoplanet search \cite{shallue2018identifying}, particle physics~\cite{radovic2018machine},  cancer detection~\cite{ribli2018detecting}, protein folding \cite{senior2020improved}, drug design~\cite{jing2018deep}, recent use has many times demonstrated super-human scientific abilities. As potentially every field can benefit one way or another of these revolutionary tools, it is natural to vigorously explore their reach within physics laboratories, which are prolific data sources. 
Among laboratory setups, many provide more or less direct insights into very relatable phenomena, like weather forecast, wildfire management, or seismicity. In the latter case, a wide array of analog systems, either closely~\cite{cryst9110582} or remotely~\cite{PhysRevLett.115.025503} related, reproduce essentially every feature of natural earthquakes. Beyond the obvious public interest, seismicity is thus a very good candidate for such program. 

Despite initial skepticism \cite{geller1997earthquakes}, seismic, or "crackling noise" prediction has received sizable historical attention in the geophysical context \cite{crampin_earthquakes_2010}, in particular regarding the renewed interest for self-organized criticality \cite{bonamy_crackling_2008-1, varotsos_self-organized_2020}, which has demonstrated some amount of predictability. This question has also been raised in other rather disparate situations, for example predicting the crackling of wood under compression \cite{viitanen_predicting_2017}, the avalanches in the archetypical sandpile problem \cite{ramos_avalanche_2009}, or metallic glasses \cite{antonaglia_bulk_2014, sun_plasticity_2010} with similar conclusions. 

However, the past five years have seen a booming development of machine-learning approaches to noisy time-series forecasting, often with impressive success, for example for option pricing~\cite{madhu2021comparative}, or epidemic propagation \cite{wu_deep_2020}. Of course, the geophysics community has been an early adopter of these approaches \cite{panakkat_neural_2007}, with many teams now working on earthquake prediction in the lab \cite{rouet-leduc_machine_2017, johnson_laboratory_2021}  and in the field \cite{asim_earthquake_2017, mohankumar_study_2018, mohankumar_study_2018-1, corbi_machine_2019, huang2018large, rouet-leduc_probing_2020}. 

One could argue that the main underlying societal stake of such prediction is its actual use in terms of population protection. However, to our knowledge, the various approaches listed above rarely put their inference results in this perspective. In practice, even if a given predictor achieves a reasonable accuracy, it remains unclear how to take advantage of this edge in a practical situation. In a sense, pure forecasting is meaningless from a political point of view, as it does not point to specific actions to be taken given the predictions. This remark deeply questions the very notion of prediction accuracy: it is highly dependent on the quantity to be predicted, and thus only carries a purely relative meaning, mostly forbidding any form of comparison, in particular from one publication to another, when the targets are distinct. 

Here we propound a methodology to overcome this challenge, based on the idea that the accuracy in the prediction of a specific quantity should be evaluated in the light of how informative it is to the associated risk management. First, we introduce a range of scalar targets to be predicted from analogue seismic signals emitted during the experimental deformation of knitted fabric, and train a generic neural network at these predictions. Then we define a risk-based metric balancing the impact of political decisions in an imaginary city (KnitCity) living on our system: evacuating the city results in a given social cost, but failing to evacuate might result in casualties. Finally, we train an agent to design a policy by minimizing this penalty over time through reinforcement learning, while having access only to the predictions of the different forecasters. This bias-free, autonomous decision making process allows to quantitatively compare the various targets and associated parameters, and to give physical meaning to the notion of accuracy within this context. Finally, we discuss the resulting ranking in terms of past and future time scales, assessing in particular the relevance of the initial forecasting endeavour.

\section{Experimental study: a relevant analogous system} 

\subsection{Experimental setup and data}
In order to gather time series of seismic-like activity, we collect experimental data while mechanically cycling a nylon thread, loose Stokinette knit, which was recently shown~\cite{poincloux_crackling_2018} to display such rich response. We use a setup similar to that reported in ~\cite{poincloux_geometry_2018}. The 83x83 stitch knitted fabric is clamped on its upper and lower rows by means of screws holding each stitch individually, imposing a constant spacing between them along the corresponding rows. Starting from an initial configuration with  height $L_i=170\,\textrm{mm}$ and width $L_c=360\,\textrm{mm}$, we perform a tensile test using an Instron$^{\mbox{\scriptsize{\textregistered}}}$ (model 5965) dynamometer mounted with a $50\,\textrm{N}$ load cell. It consists in varying cyclically the elongation $L$ of the fabric between the jaws, between $L_i$ and $L_f=220\,\textrm{mm}$. 

The mechanical response - the force signal - is recorded at high frequency (25 Hz) during the stretching phase on a shorter elongation range, between $L_m=210\,\textrm{mm}$ and $L_{f}$. 
To approach the quasi-static deformation limit in the interval $[L_m, L_f]$, we impose a constant loading speed and set it at a small value $v=5\,\mu \textrm{m/s}$. To reduce the duration of the experiment, we fix $v=5\,\textrm{mm/s}$ outside this measurement window. 

\begin{figure}[!h]
    \centering
        \includegraphics[width=0.39\textwidth]{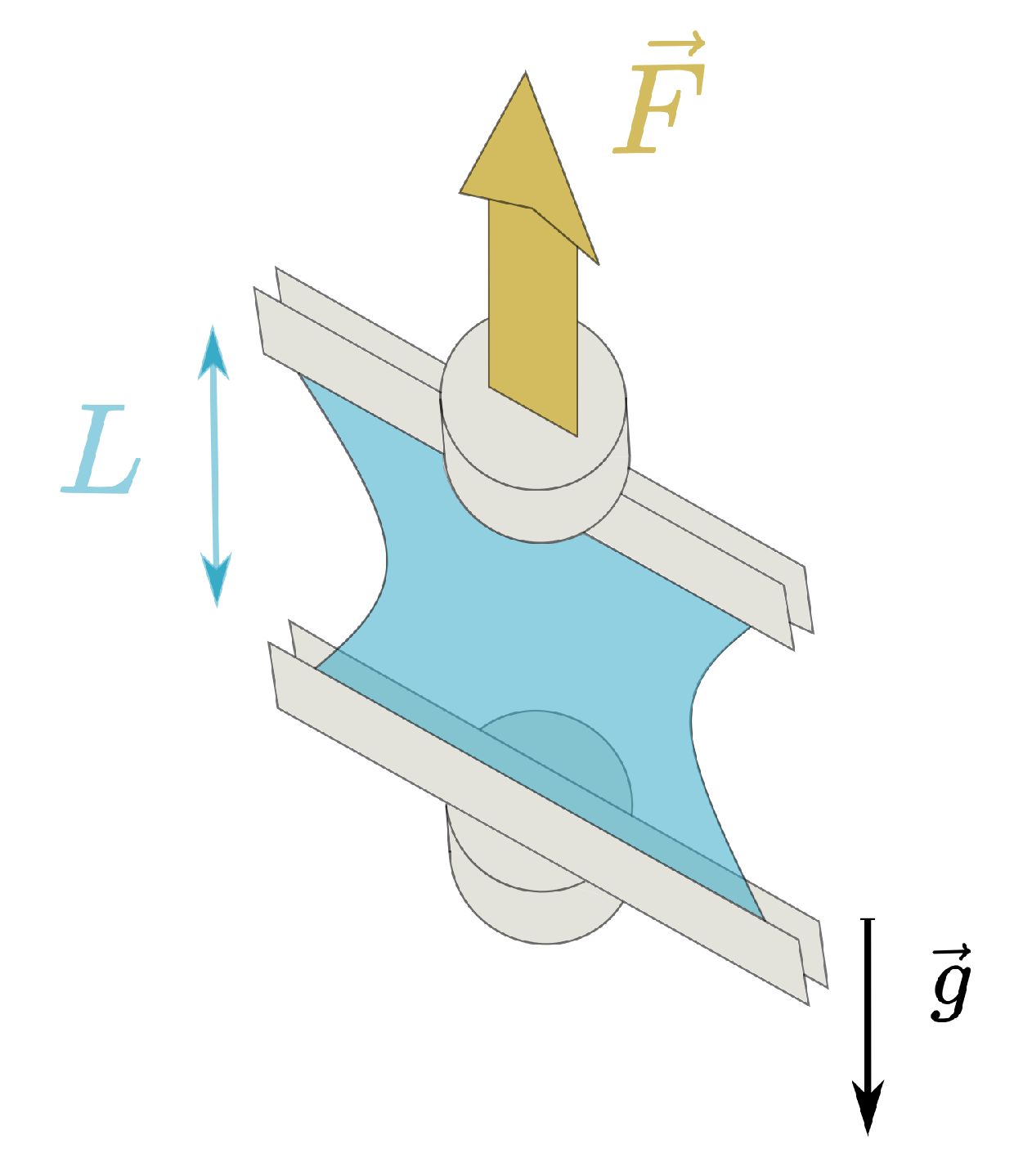}
        \caption{Color. Experimental setup: a clamped knitted fabric is set into an unixial traction device. We record both the tensile force and images of the fabric upon stretching.} \label{set_up}
\end{figure}

\begin{figure}[!h]
    \centering
        \includegraphics[width=8.6cm]{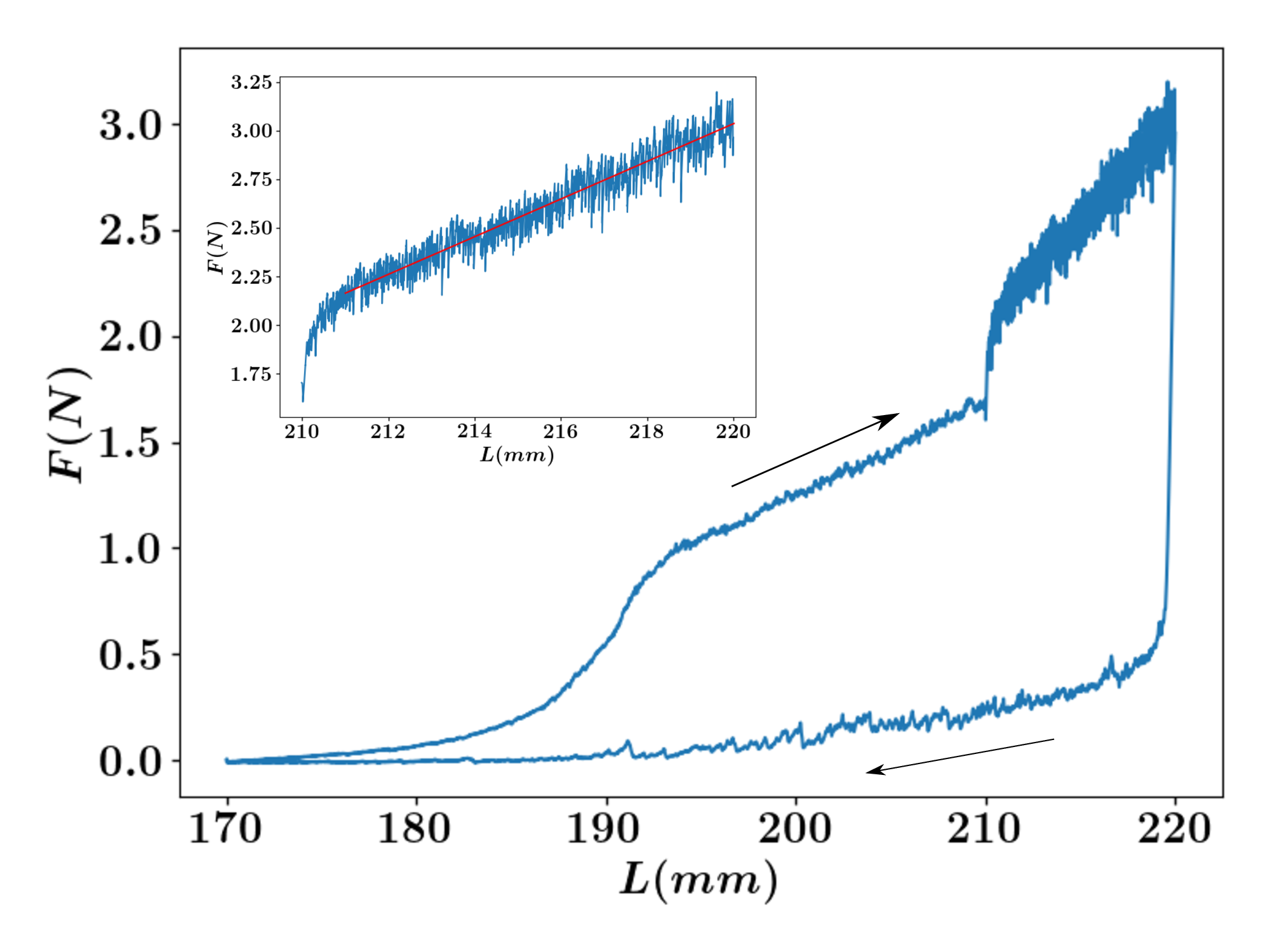}
        \caption{Color. Mechanical response of the fabric for one complete cycle of elongation followed by a compression between $L_{i}$ and $L_{f}$. The knitted fabric becomes entirely 2D around $L = 195 mm$. The slow velocity part  between 210 and 220 mm is zoomed in the upper left corner, and the affine response is highlighted in red.} 
        \label{force_et_affine_elastic}
\end{figure}

This protocol yields a value of the force every 0.04s, with a cycle duration of 2000 s, leading to 50000 points of mechanical response per cycle.  We collected data from 20 experiments of approximately 30 cycles each, which corresponds to roughly to 25 millions force points. The raw data for one cycle is shown in Fig.~\ref{force_et_affine_elastic}. 

\subsection{Data preprocessing}
Upon stretching, the mechanical response of the fabric displays large fluctuations around an affine response that is removed by a linear detrending.

The amplitude of these fluctuations is however non-stationary over the cycles, signalling the wear of the fabric over time. In order to build a robust normalization of this signal, we first remove the temporal mean of the data and scale it by its standard deviation. However, these quantities are different from on cycle to the next. Furthermore, within a given cycle, we observe that the amplitude of the fluctuations also increases linearly with the extension, presumably because the fabric has stored more potential energy. We thus apply a second such normalization, where the statistics are collected over a sliding time window over all cycles, which achieves a signal $f$ with a stationary distribution.

\subsection{Event statistics}

\begin{figure}[!h]
    \centering
        \includegraphics[width=8.6cm]{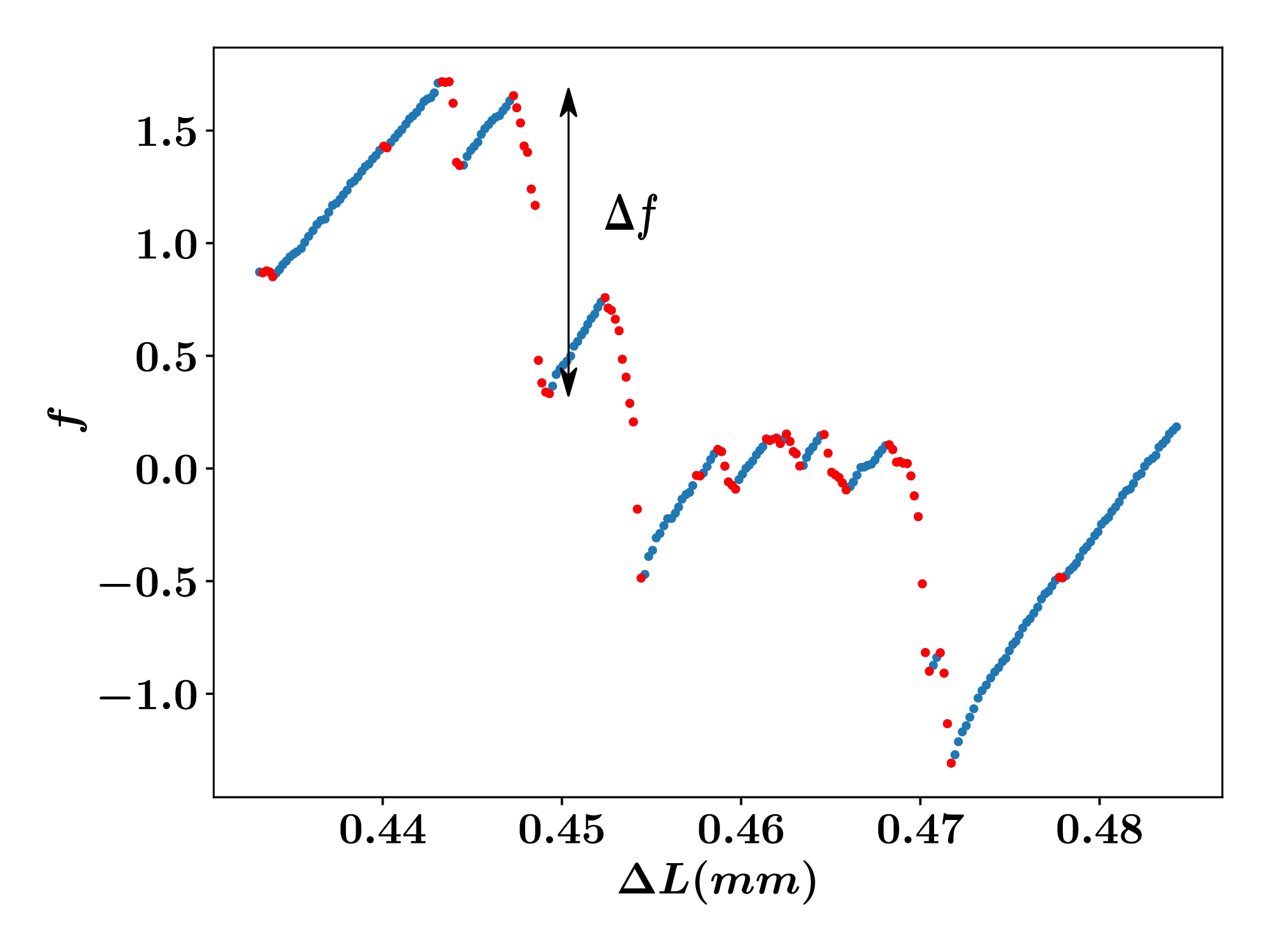}
        \caption{Color. Zoom on the normalized fluctuation signal displaying "stick" phases in blue, and "slip" phases in red. A scalar event $\Delta f$ is defined by the amplitude of "slip" phases.} \label{zoom_f}
\end{figure}

\noindent\textbf{Characterization of events --} Upon closer inspection, the normalized fluctuations $f$ exhibit a complex, "stick-slip"-like behavior, with linear loading phases interrupted by abrupt drops, as can be seen in Fig.~\ref{zoom_f}. These force drops, or "slip-events", were shown in \cite{poincloux_crackling_2018} to be concomitant and correlated in amplitude with time-intermittent, spatially extended fault-like plastic events. 

\begin{figure}[!h]
    \centering
        \includegraphics[width=8.6cm]{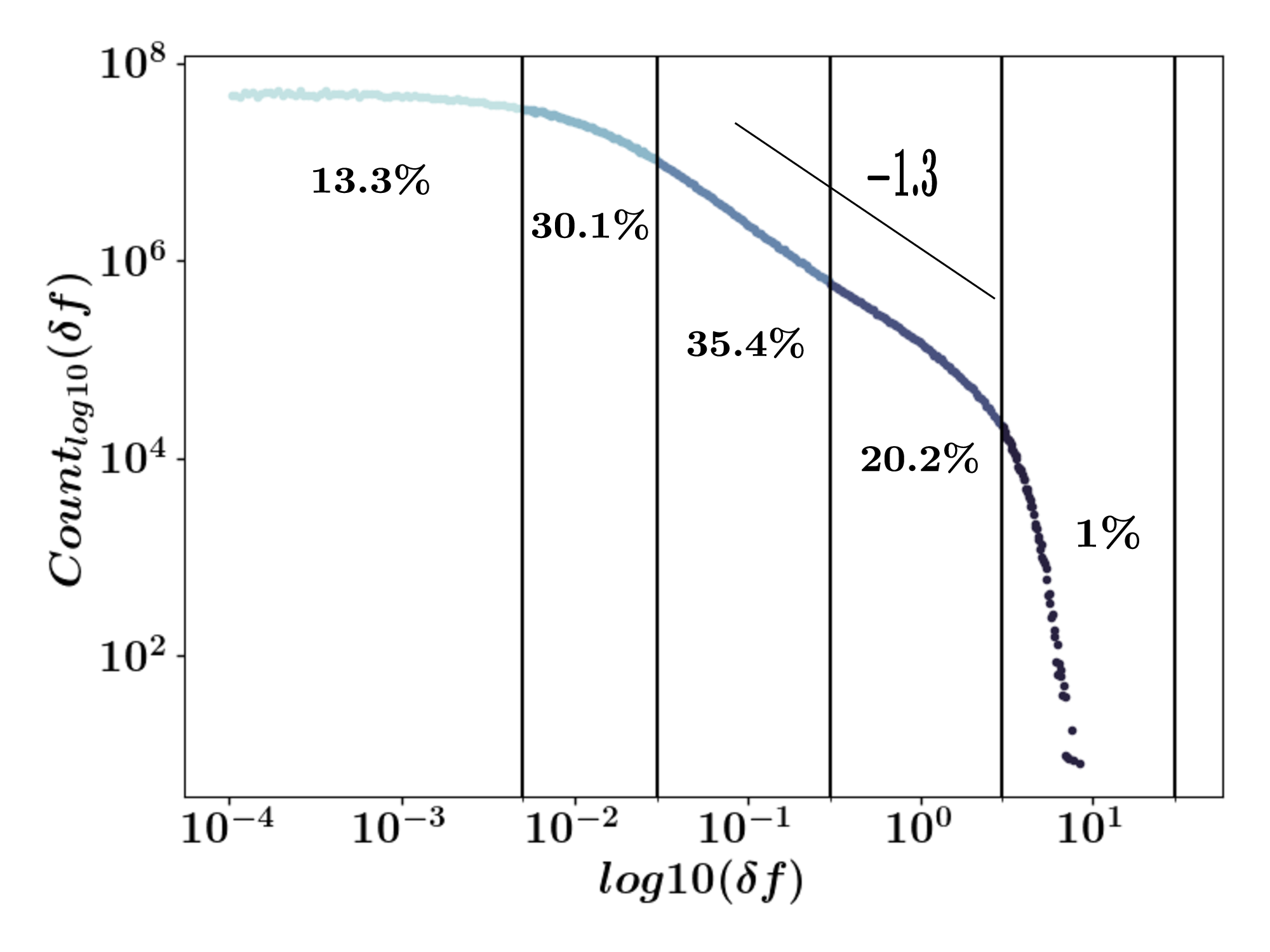}
        \caption{Color. Number count distribution function of force drops amplitudes in the normalized force fluctuations for non-zero $\delta f$'s. It shows a power law regime with exponent $-1.3$. The division into classes is based on the decimal magnitude of the force drops. The figures correspond to the relative proportion of events in each class. If looking on all $\delta f$ signal including the zeros, the proportions instead read $\left[ 93.7\%, 2.19\%, 2.58\%, 1.45\%, 0.08\% \right]$. }
        \label{stat}
\end{figure}

\noindent\textbf{Gutemberg-Richter statistics --} We are interested in these drops in the normalized loading force, as they are the equivalent of the seismic "quakes" that threaten populations, and as such are at the heart of our prediction endeavor. We therefore define the quantity $\delta f$ at each time step $t$ by:
\begin{equation}
\delta f(t) = \left\{
    \begin{array}{ll}
        \Delta f & \mbox{if } t \mbox{ is the beginning of a drop} \\
        0 & \mbox{else.}
    \end{array}
\right.
\end{equation} 
where we characterise each drop by its amplitude $\Delta f$, see Fig.~\ref{zoom_f}. Following this construction, more than 93\% of $\delta f$'s values are zeros. The events themselves, corresponding to non-zero values in $\delta f$, are roughly power-law distributed, as shown in Fig.~\ref{stat}. More precisely, the event distribution displays three different regimes : noisy structureless events of low amplitude, a scale-invariant regime that extends over slightly more than 3 decades with an exponent of -1.3, and an exponential cut-off for very large and rare events. This distribution is characteristic of the avalanching dynamics found in seismic activity \cite{ramos_avalanche_2009} and references within. 

This long-tailed statistics suggests to define classes of event severity in a logarithmic scale. As an illustrative example, we will pick five classes centered on the decimal magnitudes in the event number count distribution, as shown in Fig.~\ref{pdf_df_prop}. Class zero events correspond to noise, extended to a small event class 1. Then medium-sized events (class 2) and large events (class 3) potentially qualify as non-severe and severe respectively, while class 4 corresponds to very rare, catastrophic quakes. 

\section{Supervised machine learning}
In this section we describe in details the machine learning procedure we designed in order to predict various notions of danger imminence based on past measurements. We do time series forecasting using an artificial neural network architecture called ResNet Neural Network (NN), which is a well-known feature extractor based on a large number of convolutional layers with residual connexions~\cite{he2016deep}. 

\subsection{Building the labels}
Because most of the $\delta f$'s are zeros, the NN would only learn to predict zero without a proper rebalancing of the dataset. On the other hand, using a naively rebalanced dataset might lead to an over prediction of extreme events. We deal with this issue by introducing a future horizon window of size $\tau$ (in time steps) and predicting the severity of future events within that window. Such an aggregation of future events has indeed the automatic effect of partially rebalancing the dataset in a meaningful way. We have considered three different ways for aggregating the events in the near future: 

\begin{itemize}
	\item Target 1 (T1): 
\begin{equation}
T_1(t) = \max\limits_{\left[t, t+\tau \right]} \delta f(t)
\label{T1}
\end{equation}
This target focuses only on the largest event in the near future discarding the temporal information,
	\item Target 2 (T2):
	\begin{equation}
T_2(t) = \Sigma_{t}^{t + \tau} \delta f(t)
\label{T2}
\end{equation}  
aggregates the amplitudes of events in the near future, 
	\item Target 3 (T3): 
	\begin{equation}
T_3(t) = \Sigma_{t' = t}^{t' = t + \tau} e^{\frac{-(t'-t)}{(\tau /3)}} \delta f(t)
\label{T3}
\end{equation}
 Thanks to exponentially decaying weights, this target integrates information on both amplitudes of events and their time of  occurrence. 
\end{itemize}

\begin{figure}[!h]
    \centering
        \includegraphics[width=8.6cm]{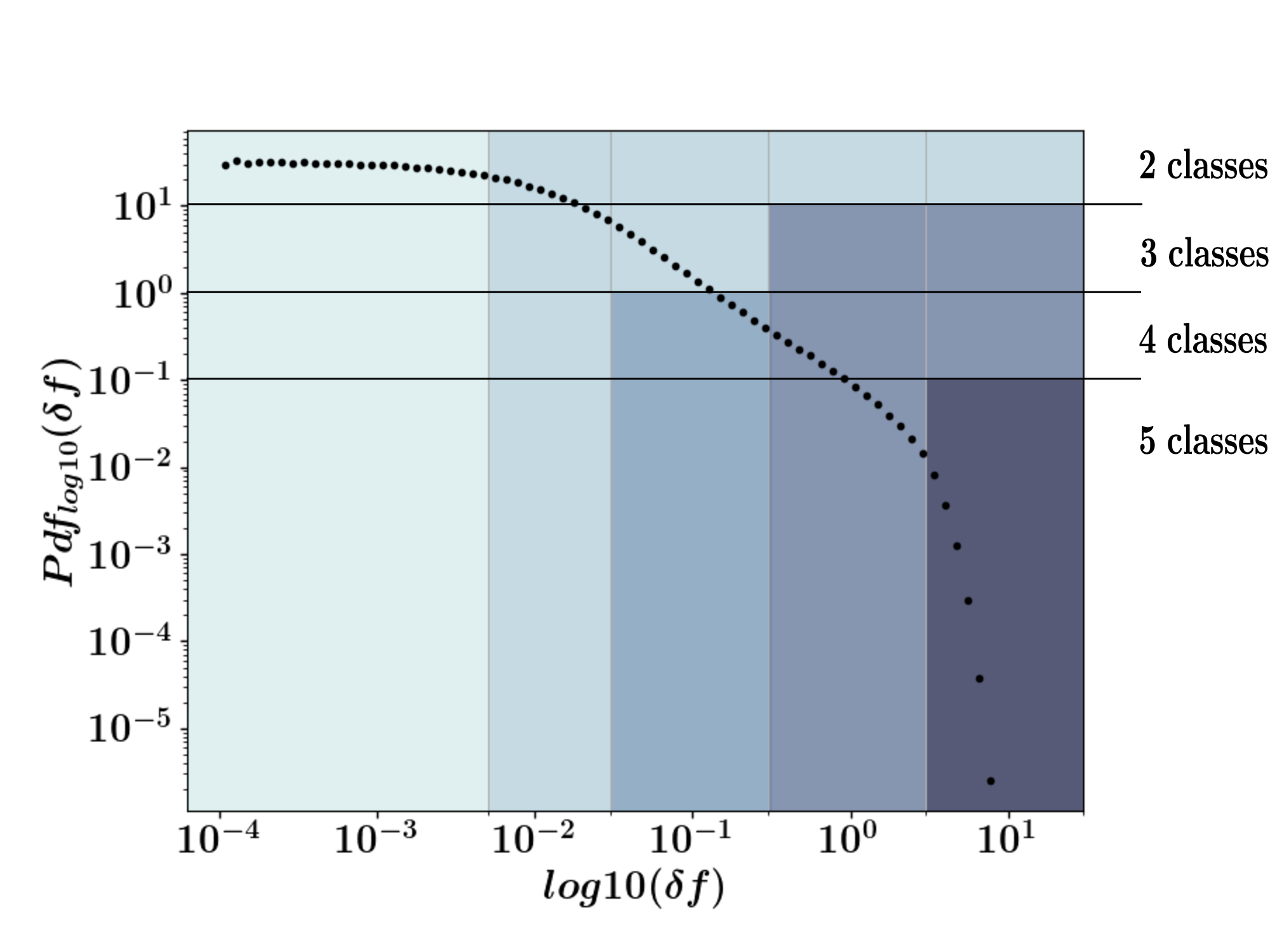}
        \caption{Color. Mapping of the target values (T1, T2, or T3) as defined in the text into N = 2, 3, 4, or 5 classes given the thresholds obtained using the Gutemberg-Richter statistics. In all cases, class 0 is the noise: eg. if N = 2 (uppermost band), targets are labeled as 0 if the scalar value is less than the threshold, and 1 otherwise. Then, by increasing the number of mapping classes we get more refined labels for the targets.
        } \label{pdf_df_prop}
\end{figure}

These scalar targets are then mapped to discrete labels, denoted Y(t) in the following, with value from 0 to N-1 corresponding to $N$ classes matching the events statistics discussed earlier - see Fig. \ref{pdf_df_prop} for details. The training dataset is then balanced with respect to these labels prior to training. We have tested different choices of $\tau = 20, 40, 60$ time steps. These values are chosen because they roughly coincide with the average time delay between events from class 1 to 3.

Note that our approach is somewhat hybrid (via the use of $\tau$) with respect to other choices of the literature, where, eg. the prediction task focused on either:
\begin{itemize}
    \item time to failure: predict the time before next event of given class as in \cite{rouet-leduc_machine_2017}
    \item future amplitude: predict the amplitude of next event regardless of when it happens.
\end{itemize}

\subsection{Inputs and model architecture}
The input of the NN is built by stacking together three 1D arrays:
\begin{itemize}
    \item $\left(f(t-n-1), \ldots, f(t-1)\right)$, 
    \item $\left(\delta f(t-n-1), \ldots, \delta f(t-1)\right)$, 
    \item $\left(Y(t-n-\tau -1), \ldots, Y(t-\tau-1)\right)$.
\end{itemize} 
Here, $n$ is a past horizon that we set to 256 time steps. We feed the neural network with both the past raw data, the past amplitude drops, and the correct past targets (with an appropriate shift $-\tau$ so that the input data contains no information about the future). 

One may object that $\delta f$ contains some future data because its value, when is non zero, corresponds to the amplitude of an event that is just beginning. However, this does not introduce future biases because what we try to predict is $T_i(\delta f)$, ie. the upcoming delta's, and not the force signal itself. We verified this claim by simply removing that data from the input, which resulted in similar performance.

We investigated different NN architectures and found that convolutionnal-based NNs achieved the best results. We will thus only report the results obtained with a ResNet18 with 1D convolutions on the 3 input channels. We used AdaDelta as optimizer with an initial learning rate of $0.01$ and a learning rate decay of $0.5$ every 50 epochs for 300-epoch runs. The model also includes weights decay set to $1.10^{-4}$.

The whole dataset of 27M sequences was split in 22M of them for training, 2.5M for the validation, and 2.5M for testing. The training sequences are then subsampled in order to enforce equiprobability of the associated labels. The test set is however not subsampled since we want to probe our model's abilities on realistic data. 

\subsection{Results}
We trained and evaluated models for the three different targets T1, T2, T3, the three future window $\tau = 20, 40, 60$, and also varying the number of classes N from 2 to 5, resulting in 36 models in total. 

\begin{figure}[!h]
    \centering
        \includegraphics[width=8.6cm]{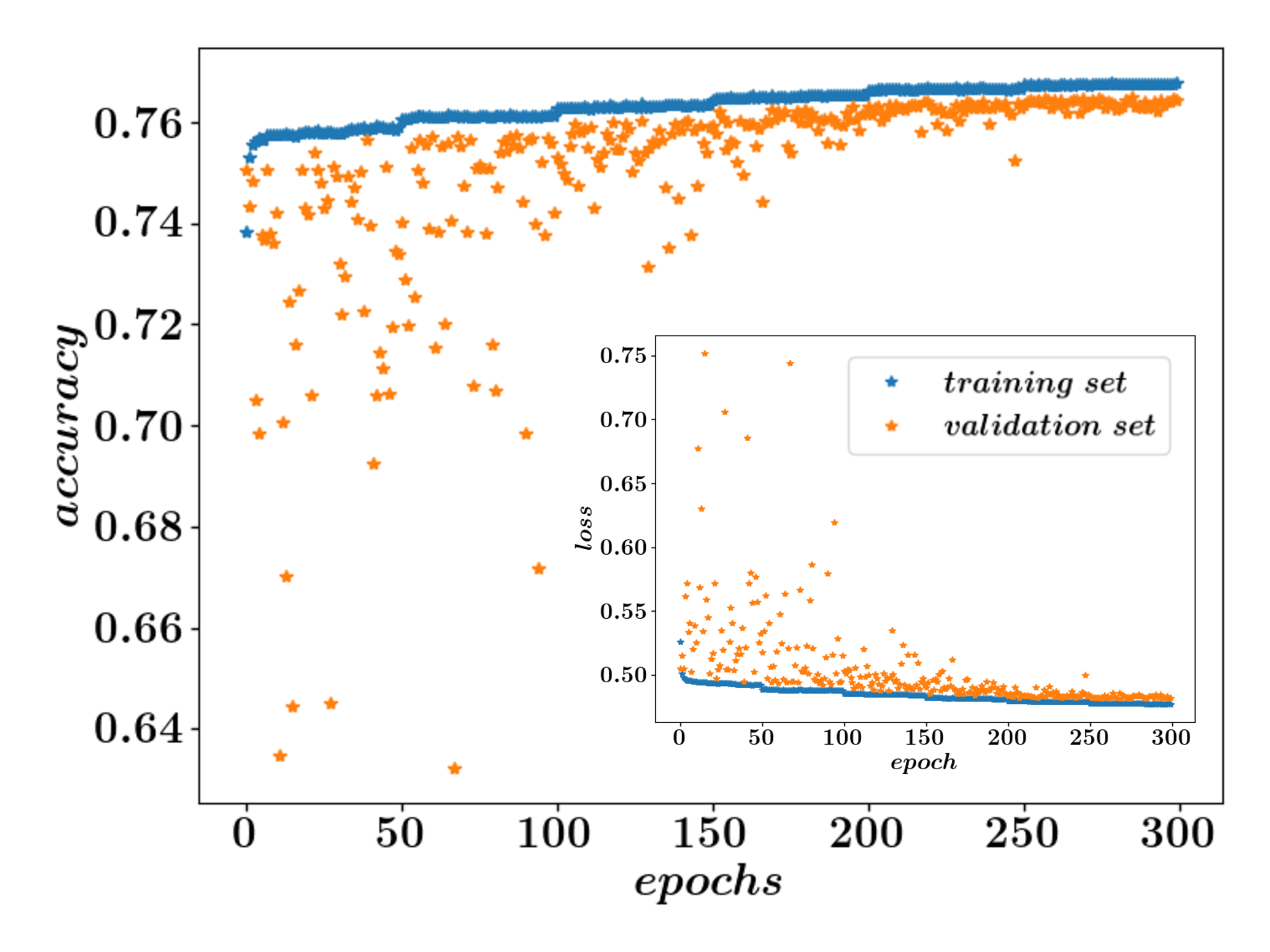}
        \caption{Color. Convergence of accuracy over 300 epochs of the learning phase for a model trained with target $T3$ and $\tau = 60$ classified over $N= 2$ classes.} \label{acc_loss}
\end{figure}

One example of learning progress is displayed in Fig.~\ref{acc_loss} for $N=2$ classes. In that case, the accuracy for both the training and validation set reaches 75\%, while the accuracy on the test set is around 74\%. This two classes model is however not very helpful in order to track our ability to predict extreme events only. But upon increasing $N$, the meaning of the accuracy value increasingly fails to capture that ability, due in particular to the proportion of events in each class being highly imbalanced. 

\begin{figure}[!h]
    \centering
        \includegraphics[width=8.6cm]{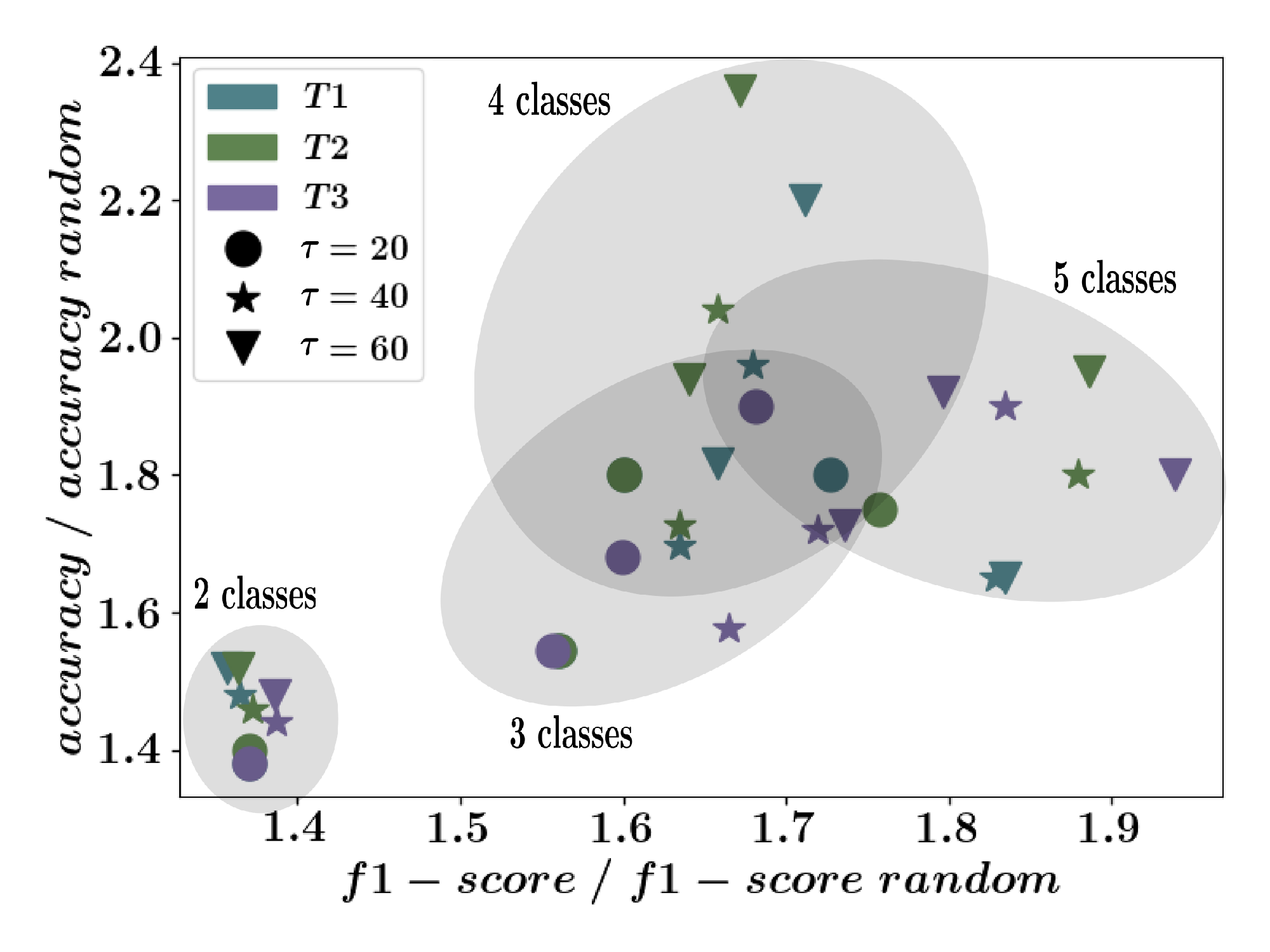}
        \caption{Color. Accuracy and F1-score for all models with respect to a random prediction.} \label{standart_acc_rsc}
\end{figure}

We thus also report the F1-score, which is an harmonic mean of recall, and precision (see appendix B for definitions), two additional and complementary metrics~\cite{mallouhy_major_2019}.
 Both these indicators, displayed in Fig.~\ref{standart_acc_rsc}, are normalized by the value they would assume with a random prediction. This figure shows that the NN is indeed able to learn and generalize from the training set, and we learn that increasing the number of classes or the future horizon $\tau$ typically leads to better performance. The target $T2$, which is simply the sum of the slip amplitudes in the future horizon, also seems to be mildly favored.

\begin{figure}[!h]
    \centering
        \includegraphics[width=8.6cm]{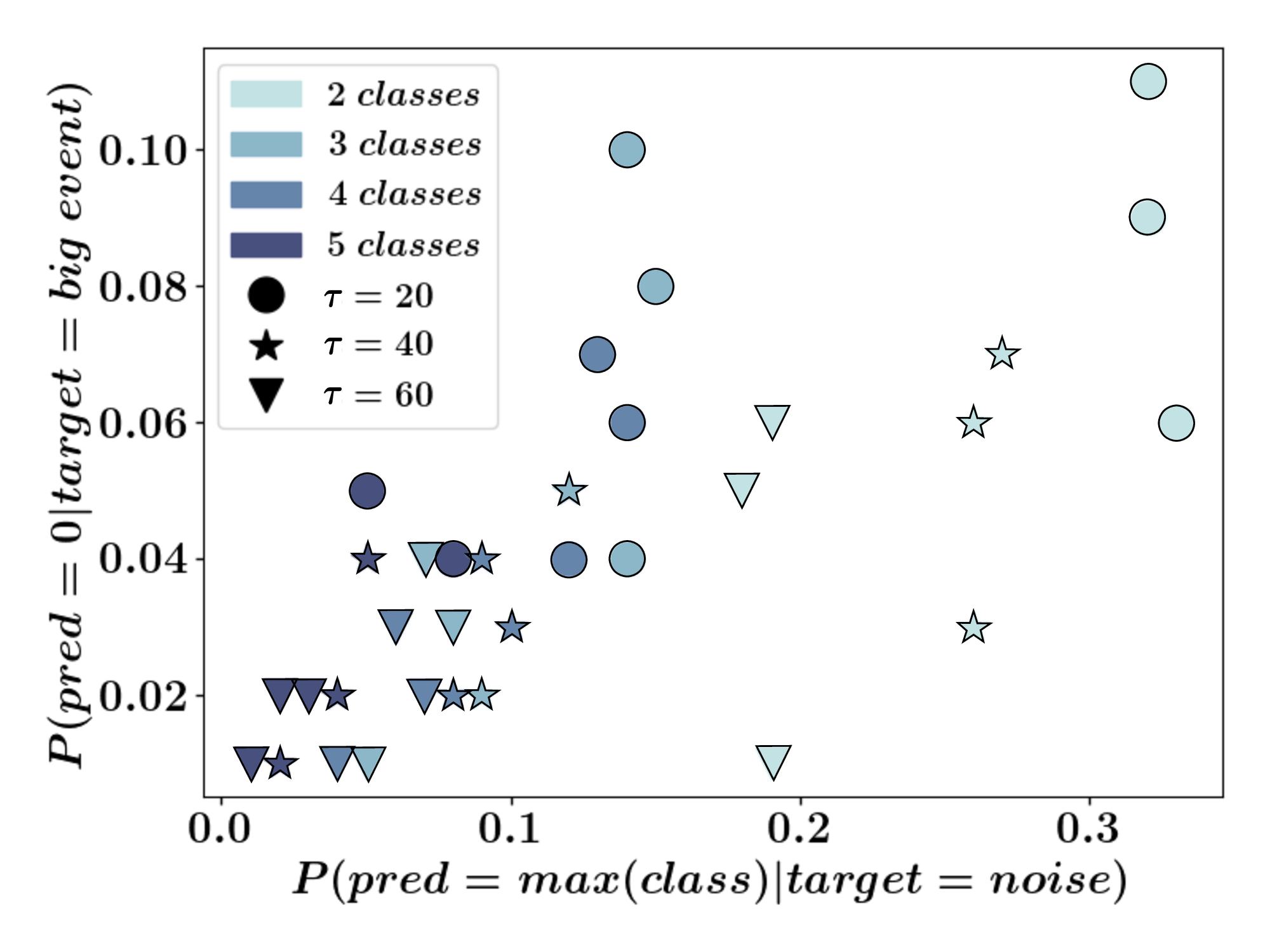}
        \caption{Color. Learning performance in the false positive (x-axis) and false-negative (y-axis) space between noise and extreme events only. Dividing the signal in more classes clearly decreases both rates. 
        } \label{risk_vs_cost}
\end{figure}

This result is confirmed by the direct study of false positive and false negative rates restricted to the two edge classes, noise versus extreme events only, as seen in Fig.~\ref{risk_vs_cost}. We see there that increasing the number of event classes greatly reduces both the risk of predicting low class events when a major event is in fact going to happen, and the risk of overestimating the danger, potentially leading to needless countermeasures. However, the interpretation of the results becomes increasingly complicated.  

Furthermore, although these false positive and negative rates may seem very low (around 1\% for the best model in the bottom left corner of Fig.~\ref{risk_vs_cost}), this is still far from being perfect because extreme events have themselves an even lower occurrence rate (around ten times smaller). 

As stated in the introduction, these predictions are not handy to read or use, and a quantitative assessment of their differential quality is still lacking at this point.One way around these difficulties consists in defining exactly what we mean by useful prediction: within the georisk context, a prediction is useful if it allows policy-makers to make better informed decisions, in order in particular to save more lives while sparing the local economic activity as much as possible.  
In the next section we thus describe a second layer of (reinforcement) learning meant to design a risk-management policy based on the predictions of this first layer. The idea is that upon inspection of the past predictions of the model, we may on the one hand be able to compare the different models in terms of a single risk-based metric, and on the second hand be able to improve the decision of evacuating or not the city with respect to some naive policy by taking advantage of the \textit{dynamics} of the past predictions of event severity. 

\section{Game-theoretical framework}

\subsection{Introducting KnitCity}
In this section, we introduce KnitCity, a virtual city that is subjected to "knitquakes" based on the crackling dynamics extracted from the mechanical response of the knitted fabric. KnitCity laboratory keeps track of all past seismic data, and has access to all past predictions of the Neural Network designed previously. 

At each time step, the mayor must decide whether to evacuate the city or not. Such a decision is a trade-off between the risk of exposing citizens to catastrophic quakes resulting in a severe human loss and the risk of evacuating the city for nothing, to which we may associate a "social cost", related to the corresponding economic loss for example. Such a social cost will be denoted $\lambda$ for each day spent out of the city. 

On the other hand, we model the human cost with the following weights corresponding respectively to events from class 0 to 4:
\begin{equation}
\textrm{damage} = \mu*\left[0, 0, 0, 1, 10 \right]
\label{damagedef}
\end{equation}
where $\mu$ is a free parameter. Such a modelling choice reflects the fact that for actual earthquakes, there is a threshold below which human cost most often vanishes. Also, increasing by an order of magnitude the damages caused by class 4 events with respect to class 3 events reflects the fact that amplitude drops, which embody the events severity, increase exponentially with class label.
Since the ratio $\mu/\lambda$ is the only parameter that drives the trade-off between casualties and time out, $\lambda$ can be arbitrarily set to 1. There are then two limiting behaviors: if $\mu$ goes to infinity, the optimal choice is to always stay out of the city, and to always stay in if $\mu$ goes to zero. Although the precise value of $\mu$ cannot be estimated a priori since it is based on social considerations, it must lie somewhere in between these two limits for otherwise there would be nothing to learn. 

Using Eq.~(\ref{damagedef}) and the event statistics, we may estimate typical values of $\mu$ by requiring that a random policy that evacuates the city $p$ percent of the time yields a social cost of the same order of magnitude as the human cost. Values of $p$ around $0.2$ or less (see results section below) led to the order of magnitude: $\mu = \mathcal{O}\left(20\right)$.

A decision policy $\pi$ can be evaluated by a reward given by the sum over an episode (defined by some number of contiguous time steps, see below) : 
\begin{equation}
R = \sum_{t \in episode} s(a(t)) + h(a(t), \delta f(t))
\end{equation}
where $s$  (0 or -1) is the social cost determined by action $a$ at time $t$, and $h$ is the human cost that depends on both the action and the actual event $\delta f$:
\begin{equation}
s(a)  = \left\{
    \begin{array}{ll}
    - 1 & \mbox{if action: leave}\\
       0 & \mbox{if action: stay}.
    \end{array}
\right.
\end{equation} 

\begin{equation}
h(a, \delta f)  = \left\{
    \begin{array}{ll} 0  & \mbox{if a == leave} \\
        - \textrm{damage}(\delta f)  & \mbox{if a == stay}.
    \end{array}
\right.
\end{equation} 

\begin{figure}[!h]
    \centering
        \includegraphics[width=8.6cm]{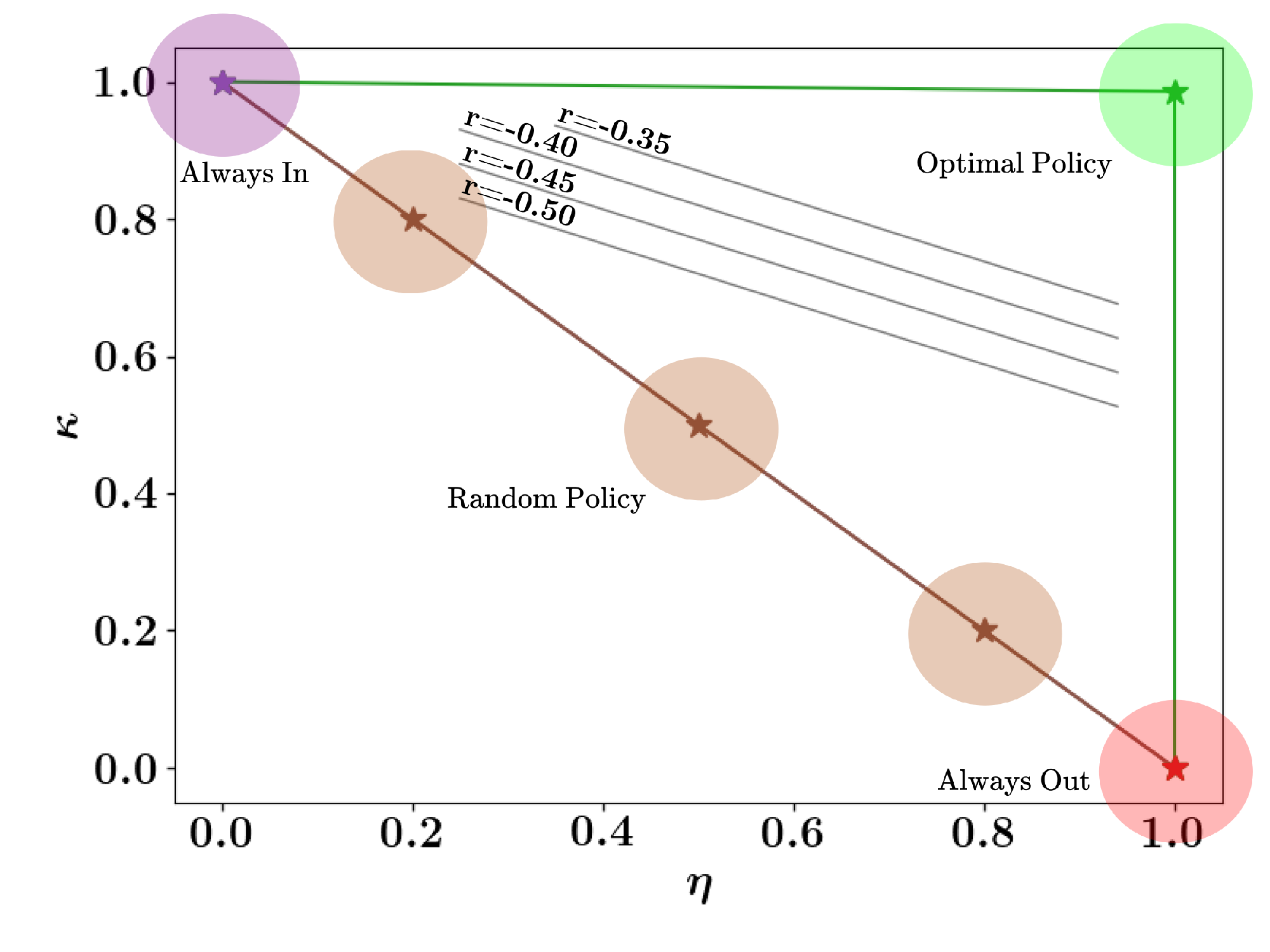}
        \caption{Color. Policy performance in the $\eta$ and $\kappa$ space. Random policies are shown as well as limiting policies $(1,0)$ or $(0,1)$ as discussed in the text. The optimal agent is in the top right corner. Also shown are some of the lines that define constant reward per step $r$. The slope of these straight lines directly depends on the chosen value of $\mu$: they have increasingly negative slope for increasing values of $\mu$. Here $\mu = 20$.} \label{extrem_policy}
\end{figure}

In the following, we shall also use two more quantities in order to represent policy performances. First we define
\begin{equation}
\eta = 1 - \frac{\mbox{actual casualties following the policy}}{\mbox{maximum casualties}}
\end{equation}
the average life saving rate and
\begin{equation}
\kappa = 1 - \frac{\mbox{number of steps where we evacuate}}{\mbox{number of steps in the episode}}
\end{equation}
the average population retention rate. We may then define two dimensional diagrams in which, for example, the policy consisting of always staying in the city has coordinates $(0, 1)$ (see Fig. \ref{extrem_policy}), while the policy consisting in always evacuating the city has coordinates $(1, 0)$ in the ($\eta, \kappa$) space. An agent that randomly evacuates the city at some rate $p$ should on average expect casualties proportional to $1-p$ and thus the set of all possible random policies must lie on average on the $y = 1-x$ straight line. Below that line are policies worse than random.

The optimal policy that consists in evacuating the city only when the damages are larger than the social cost can be evaluated by simply looking into the future: we find $\eta = 1$ and $\kappa$ also very close to 1 since dangerous events are very scarce. If we moreover add that the evacuation rate is fixed and continuously varying from 0 to 1, then we may define "an optimal frontier" as being the green straight lines in Fig. \ref{extrem_policy}. 

In that space, the actual reward by time step can also be mapped. It is easily shown to be $r =  r(\kappa, \eta) =  A_1 - \kappa - A_2 \eta$ via the previous equations, for two constants $A_1$ and $A_2$ that can be computed on the signal statistics and on $\mu$. Models can thus be ranked with respect to the average reward per time step; constant reward are decreasing straight lines in that space, see e.g. Fig. \ref{extrem_policy}.

\subsection{Training a decision making agent}
Using the NN's predictions of section 3, we first manually define a naive policy based only the last available prediction, where we evacuate at every alert beyond the threshold $\lambda$, ie. every time an event of class 3 or 4 is predicted:
\begin{equation}
\pi_{\textrm{naive}} = \left\{
    \begin{array}{ll}
       \textrm{leave} & \mbox{if } \textrm{damage(prediction)} > \lambda  \\
        \textrm{stay} & \mbox{else.}
    \end{array}
\right.
\end{equation} 
Results of such a naive policy are shown in Fig.\ref{mainresult}. The natural question is however whether we can go beyond the naive policy by leveraging the sequence of past NN's predictions through the training of a reinforcement learning agent.

We therefore trained an unsupervised model consisting in a Categorical Deep-Q-Learner \cite{bellemare_distributional_2017}  belonging to the wide class of Reinforcement Learning (RL) Agents that have achieved remarkable performance in many areas that need the time component and memory to be taken into account : from games to control, auto-pilots, etc. Such a network learns the probability distribution function of the expected rewards for a given action and a current state, and acts by sampling it. 

We use a distributional Q-function with discreet action of 51 atoms, with 2 hidden layers of 64 neurons. As optimizer we use Adam with an  $1.10^{-3}$ rate, and a discounting factor $\gamma$ of $0.95$.

We subsampled the test set of the previous section III. B in a set of 75 episodes, or "games" of 5000 time step each (such that on average, two major events of class 4 are present in each game), and validate on unseen data grouped in 50 episodes of 5000 steps. All models are then tested on 180 new unseen episodes of the same number of steps. We trained multiple RL agents that had access to a varying number of past predictions.

\subsection{Results}

\begin{figure}[!h]
    \centering
        \includegraphics[width=8.6cm]{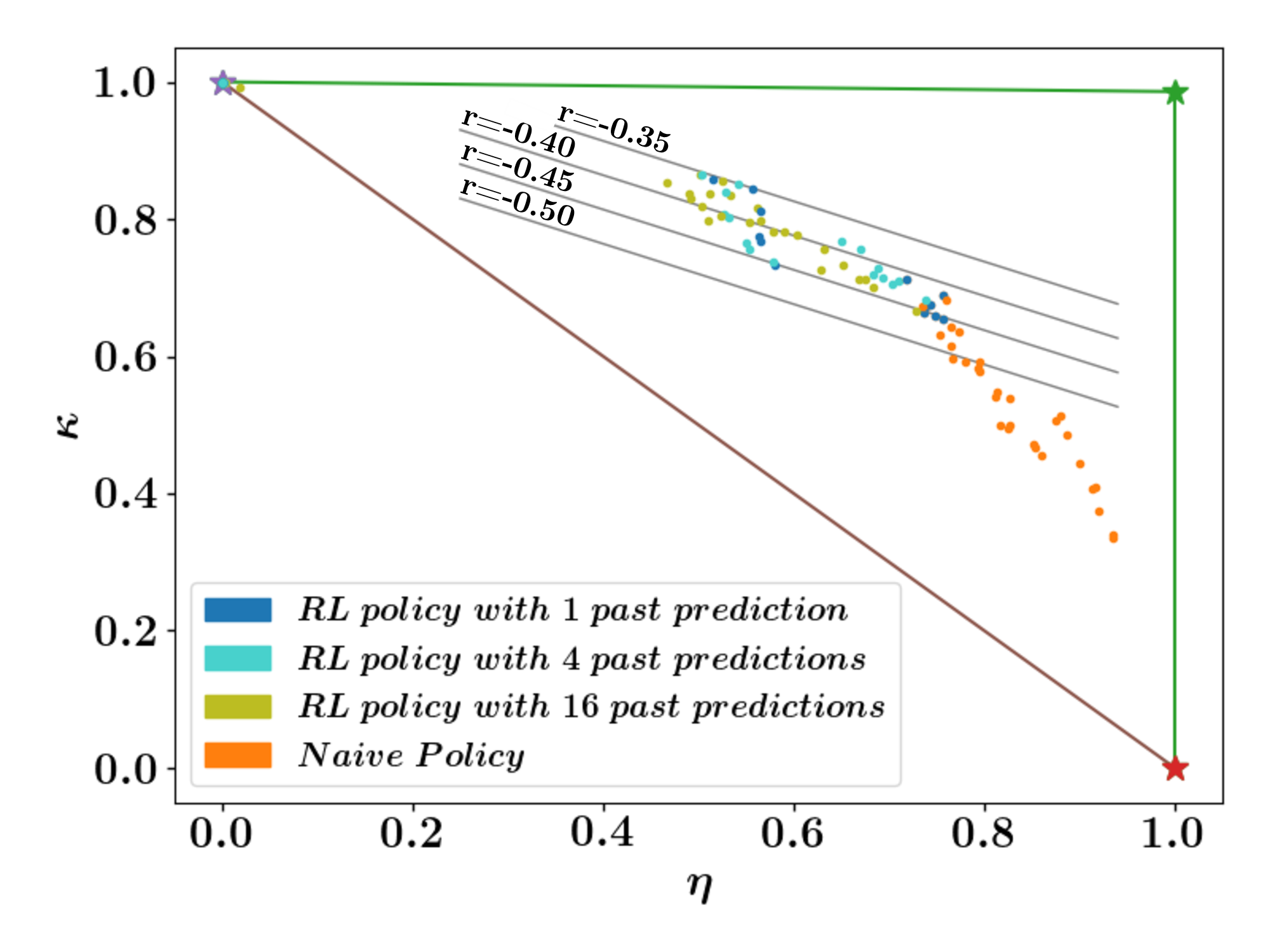}
        \caption{Color. Ranking of learned policies with an RL agent compared to the naive policy for the 36 different models trained in the previous section. Also shown are the lines of constant reward per time-step. Some of the models converged to uninteresting always-in/out policies.} \label{mainresult}
\end{figure}

We first fix $\mu$ to 20, and train the RL agent for the 36 models obtained in the previous section, with both 1, 4 and 16 past predictions made available to the RL agent. We will denote RL(1), RL(4), ... and so on these models. In terms of the average reward per step, the results in Fig.~\ref{mainresult} clearly shows that RL agents improve a lot the naive policy. Notice also that some of the models fail to converge to an interesting policy but instead converge to the "always in" policy (especially RL(1) models). 

Zooming in on the best performing models with a reward per time step $|r|$ below $0.4$, we can return a list of best models,
Irrespective of the number of past data fed to the RL agent, the best model is systematically the one that uses $N=5$ classes with target 3 (exponential decay [see Eq. (\ref{T3})]) and a future horizon of 20 or 40.

\begin{figure}[!h]
    \centering
        \includegraphics[width=8.6cm]{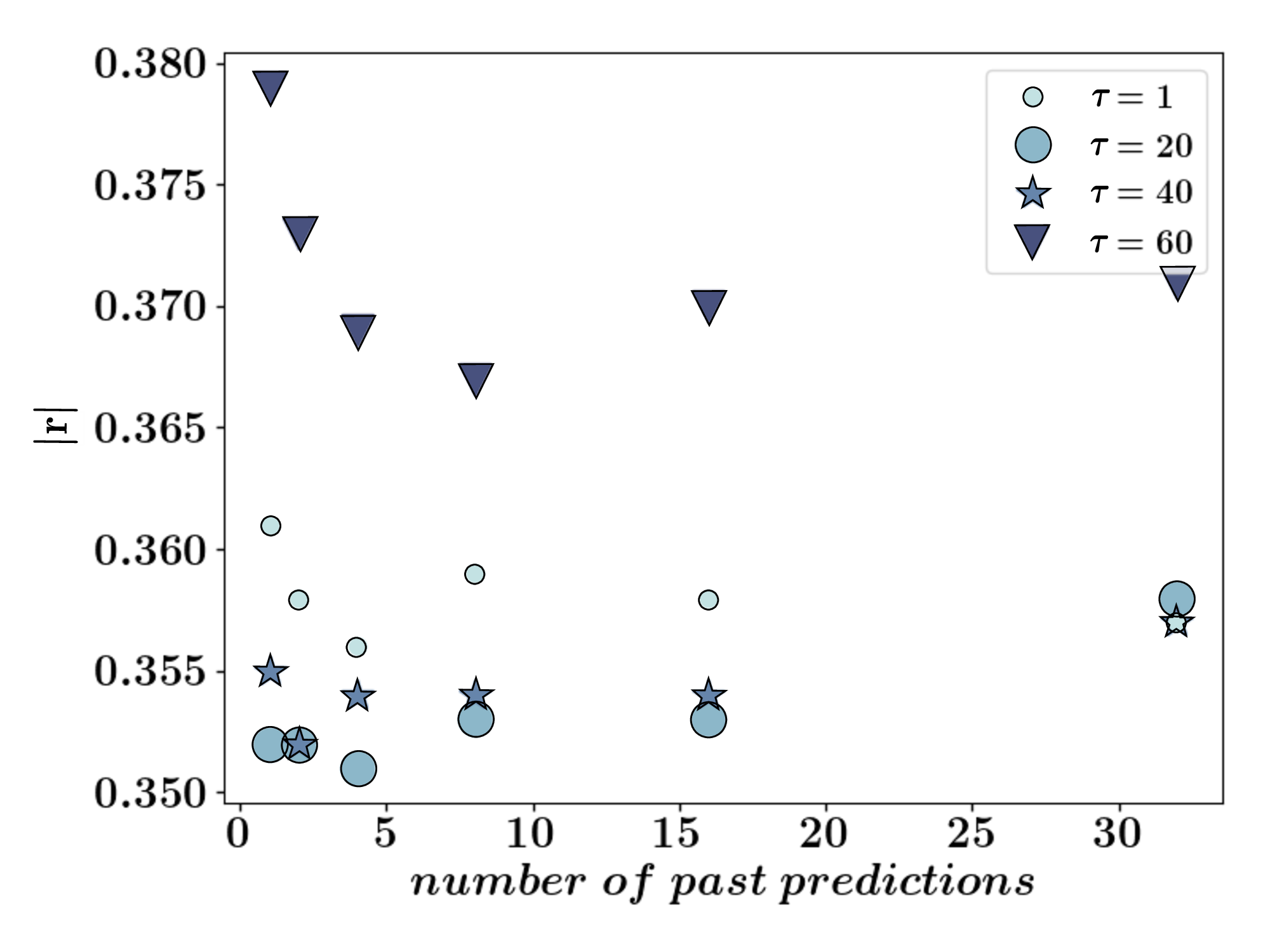}
        \caption{Color. Ranking of learned policies for the model defined with $N=5$ classes and target $T3$. The largest impact on the results comes from the size of the future horizon, with $\tau$ around 20 being clearly favored. The number of past predictions fed to the RL agent is less sensitive, although we find a best model for $n=4$ past predictions, and a slight decrease of learning performance if we feed too many of them. Here the reward per step is plotted in absolute value, and best models have the lowest $|r|$ values.} 
    \label{mainresult2}
\end{figure}

Exploring further our results, we now fix the model (target 3 and $N=5$), and vary $\tau$ in the range $(1, 20, 40, 60)$ and for $(1, 2, 4, 8, 16, 32)$ past predictions made available to the RL agent. We got the the results of Fig.~\ref{mainresult2}.

\begin{figure}[!h]
    \centering
        \includegraphics[width=8.6cm]{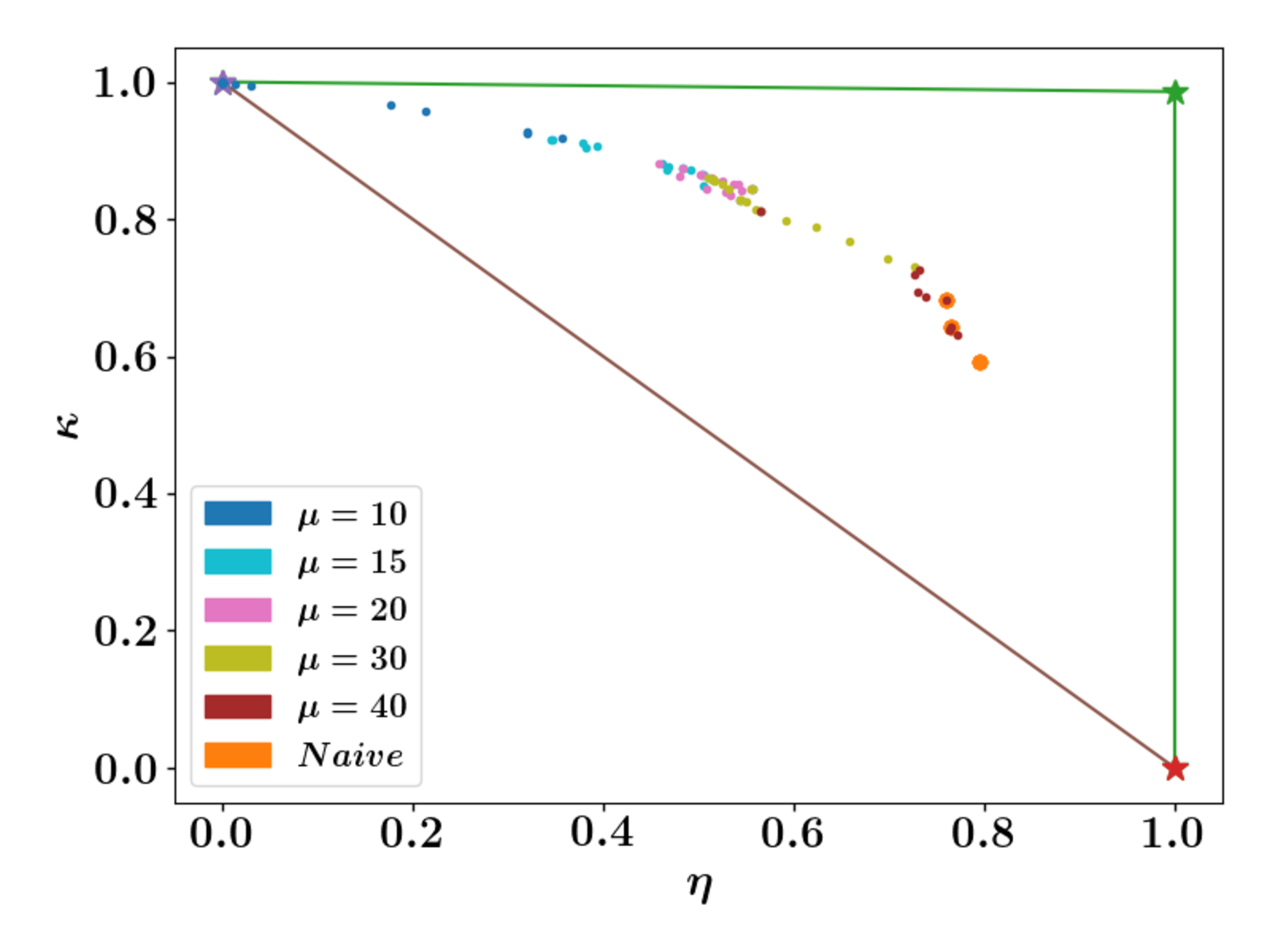}
        \caption{Color. Typical positions of learned policies in the $\eta$, $\kappa$ space when $\mu$ is varied; for models with $T_3$, $N=5$ classes and $\tau$ in the tuple $(20, 40, 60)$, and different number of past predictions.
        } \label{muvaluesresults}
\end{figure}

Finally, we report on the effect of varying the parameter $\mu$. This trend is shown in Fig. \ref{muvaluesresults}. Points here are trained models with target 3 and 5 classes, for different future horizon and different RL(n), but are grouped only by mu-values. This enables us to see quite a smooth evolution from always-in (low $\mu$) to always-out policies in accordance with our expectations.

Again, this figure emphasizes that there is no unique way to define an optimal policy when predictions are not perfect; instead we get a family of trained decision agents that are $\mu$ dependent. However the point is that the framework we introduced now provides with both quantitative and qualitative tools to explore and rank decision policies, and most importantly it enables us to compare models between one another, and even between experiments of completely different nature.

\subsection{Discussion}
In an effort to gain insights into the immediate future crackling activity in the dynamometric signal of a knitted fabric, we have established a general framework, applicable {\it de facto} to any time series prediction for which a clear motivation can be rationalized, in which we can on the one hand compare through a common rating the predictive power of the predictor, and on the other explicitly construct a step by step action plan answering the initial motivation by efficiently exploiting the predictions. This development allows in particular to rank the relevance of the various targets, past and future horizons used in neural network predictors for the risk management in a model seismic environment. Encouragingly, our analysis clearly shows that these predictors yield some valuable information about the future. 

Of course many challenges remain to be tackled within our approach. Extensive optimization of hyperparameters would certainly allow substantial refinement both in the quality of the predictor and that of the decision-making agent: our work only provides a proof of concept of the type of insights that can be obtained within the model environment of KnitCity, and was not fully fine-tuned. Building on our findings, it would be worth further investigating the role of the number of output classes, even going to the regression limit. 

Furthermore, our dataset allows for spatially resolved predictions \cite{poincloux_crackling_2018}, which would of course also be beneficial within the geoseismic context as satellite imaging enables such analysis: this approach is high on our list of future priorities. However, establishing a quantitative correspondance between the space and time scales in our system and those in the geophysical context would be required if we were to export the machinery introduced above to real-world situations, and this preoccupation will be at the heart of our future work on spatially resolved prediction.

\begin{acknowledgments}
We thank Sebastien Moulinet for his assistance in controling the experiment. We acknowledge the PaRis Artificial Intelligence Research InstitutE and Floran Krzakala for financial support. 
\\ \\
\end{acknowledgments}

\section*{Appendix A: Knitted fabric}

The fabrics were crafted using a Toyota KS858 single bed knitting machine and were made of a nylon-based mono-filament (Stroft$^{\mbox{\scriptsize{\textregistered}}}$ GTM) of diameter $d=150\,\mu\textrm{m}$ and length $25.4\,\textrm{m}$. All samples are composed of $83\times83$ stitches with an average lateral and longitudinal size of, respectively, 3.9 and 2.8 mm. The yarn Young's modulus $E\approx 5.1\,\textrm{GPa}$ was measured using a tensile test, yielding a bending modulus $B\approx 10^{-8}\, \textrm{J.m}$. 

\section*{Appendix B: Metrics}

We recall the definition of standard classification metrics for reference.
\begin{equation}
\text { accuracy }=\frac{\text{number of true predictions}}{\text{total number of elements}}
\end{equation}

\begin{equation}
\text { f1-score }=\frac{2*\text{precision}*\text{recall}}{\text{precision}+\text{recall}}
\end{equation}

\begin{equation}
\text {precision }= \frac{1}{\text{number of classes}} \sum_i precision(i) 
\end{equation}
\begin{equation}
\text {recall }=\frac{1}{\text{number of classes}} \sum_i recall(i)
\end{equation}

\begin{equation}
precision(i) = \frac{\text{number of true predicitons in class i}}{\text{number of elements predicted in class i}}
\end{equation}
\begin{equation}
recall(i)= \frac{\text{number of true predicitons in class i}}{\text{number of elements in class i}}
\end{equation}

\bibliography{refs}

\end{document}